\documentclass[conference]{IEEEtran}
\usepackage{amsmath,epsfig}
\usepackage{amsfonts}

\ifCLASSINFOpdf
\else
\fi
\begin{document}
%
\title{TP$_{sgt}$R: Neural-Symbolic Tensor Product Scene-Graph-Triplet Representation for Image Captioning}

\author{
\IEEEauthorblockN{Chiranjib Sur} %
\IEEEauthorblockA{
University of Florida\\
Gainesville, Florida 32611\\
Email: chiranjib@ufl.edu} %
}


%


\maketitle

\begin{abstract}
Image captioning can be improved if the structure of the graphical representations can be formulated with conceptual positional binding. In this work, we have introduced a novel technique for caption generation using the neural-symbolic encoding of the scene-graphs, derived from regional visual information of the images and we call it Tensor Product Scene-Graph-Triplet Representation (TP$_{sgt}$R).  While, most of the previous works concentrated on identification of the object features in images, we introduce a neuro-symbolic embedding that can embed identified relationships among different regions of the image into concrete forms, instead of relying on the model to compose for any/all combinations. These neural symbolic representation helps in better definition of the neural symbolic space for neuro-symbolic attention and can be transformed to better captions. With this approach, we introduced two novel architectures (TP$_{sgt}$R-TDBU and TP$_{sgt}$R-sTDBU) for comparison and experiment result demonstrates that our approaches outperformed the other models, and generated captions are more comprehensive and natural.  
\end{abstract}


%
\IEEEpeerreviewmaketitle

\section{Introduction}\label{sec:int}
Image captioning has gained attention due to its enormous utility in many applications relating digital media with the language world. Most of the previous works were concentrated on convolutional composition of image features (CNN) as an encoder and a recurrent neural network (RNN) like module as its decoder \cite{Karpathy2014Deep, vinyals2017show}. However, they lacked external information (such as semantics \cite{gan2017semantic}, regional details \cite{anderson2018bottom}) that could provide extra information and enhance the performance of the models. Previous works in Tensor Product Representation \cite{smolensky1990tensor, palangi2017question, chiranjib2019semantic} found TPR to be useful as comprehensive contexts for decoded words along with predicted grammatical role vector at each time step. 
In this work, we tried to exploit the positional structural features of the graphs derived from images using  the concept of Tensor Product Representation \cite{smolensky1990tensor}, which is based on coupling of context and role vectors. Previous works on Semantic TPR \cite{chiranjib2019semantic} worked on exploitation of grammar structures that could be used to predict the later words through a grammatical overview. This work introduced a novel concept of graphical embedding of the scene-graph triplets through series of orthogonal positional composition and demonstrated that the graphical overview of the regional object interactions can contribute to better caption generation. The main motivation of this work is to define neuro structural composition instead of learning weights, which limits the operational distribution of the feature space. 
\begin{figure}[h]
\centering
\includegraphics[width=.5\textwidth]{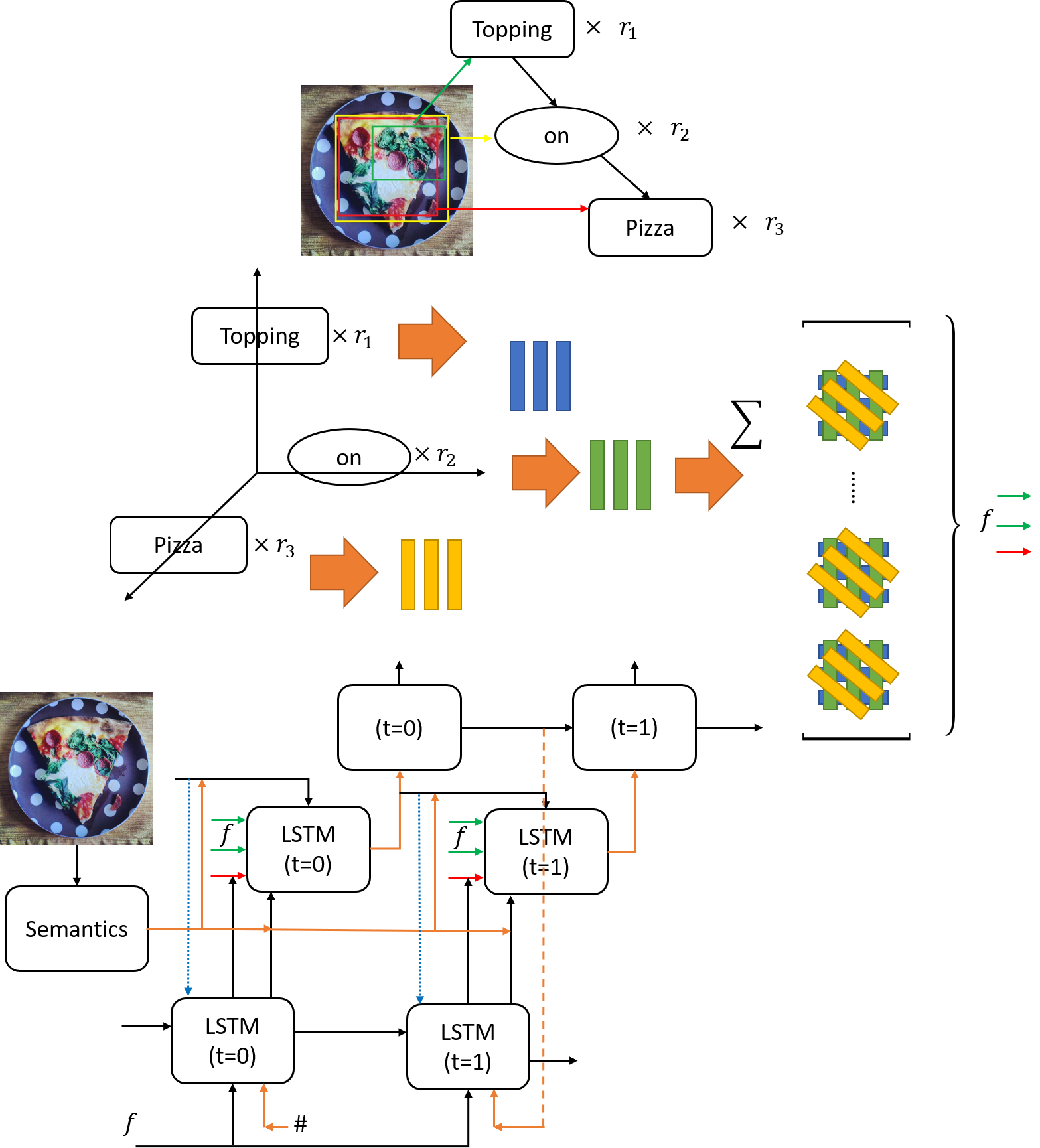}  
\caption{Architecture of Semantic Top-Down TP$_{sgt}$R Bottom-Up (TP$_{sgt}$R-sTDBU) Network. TP$_{sgt}$R-sTDBU in Lower Portion and  TP$_{sgt}$R Generator in the Upper Portion. We used Semantic Layer Information with the Language Decoder Recurrent Unit.} \label{fig:SPO-sTDBU} %
\end{figure}

The main contributions of this paper are as follows: (i) Inspired by the idea that precious words embedding based context and positional clue for later word prediction, the proposed Tensor Product Scene-Graph-Triplet Representation (TP$_{sgt}$R) model helps in better generation of the finished sentence through neuro-symbolic structuring of graphical representation. (ii) we outperformed many previous works for image captioning benchmark of MS COCO dataset. The rest of the document is arranged with revisit of existing literature in Section \ref{sec:re}, brief description of architecture and variations in Section \ref{sec:tp}, results and analysis in Section \ref{sec:res} and conclusion in Section \ref{section:discussion}.

\section{Related Works} \label{sec:re}
Previous works introduced different architectures like multi-modal visual features based language embedding like in \cite{Karpathy2014Deep}, while \cite{wang2016image} proposed a bidirectional LSTM language decoder and \cite{vinyals2017show} introduced a generative model with Inception features. Attention was introduced like regional combination emphasis \cite{lu2017knowing}, while \cite{wu2017image} utilized a high-level concept based attribute attention layer, \cite{yang1605encode} introduced a review attention for decoding. 
\cite{lu2018neural} introduced a sentence template-based approach with explicit slots correlated with different specific image regions using R-CNN objects predictors. \cite{anderson2018bottom} introduced the bottom up top down approach with emphasis on regional ResNet embedding with regional proposal network (RPN) help. Meanwhile, \cite{gan2017semantic} introduced semantic information for LSTM to be converted into sentences from images, where additional semantic concepts come from image features. Another instance of higher level attribute attention was introduced by \cite{chen2018show} where the attributes were the objects detected in the images and a separate RNN network was used for detection of these good objects from the images in a sequence that can be favorable for better caption generation. 
All these previous models were based on the co-occurrence dependencies of different object attributes and used an inference representation from these objects attributes, which prevents proper utilization of the full potential of the regional features. 
On the contrary, our approach utilized structuring of data through neuro-symbolic definition where the image features combined orthogonally to generate the representation. Like previous TPR models, these structured representations do not undergo suppression of information and emphasize on proper feature combination generation for caption generation.  The key of our architecture is that we successfully deployed a neural network to explore the structural richness of the graphical embedding of TP$_{sgt}$R and successfully outperformed many previous models. 
A direct comparison of these external feature models cannot be made with our approach as we are dependent on scene-graph generator models like \cite{li2018factorizable} for features which are trained with different set of data, but we have reported competitive performance with our approaches.

\section{Tensor Product Scene-Graph-Triplet Representation (TP$_{sgt}$R)} 
\label{sec:tp}
The graphical embedding of Scene-Graph (SG) with Positioning is regarded as Tensor Product Scene-Graph-Triplet Representation (TP$_{sgt}$R), where positioning infers spatial (role) information and features (context) are contents like Subject-Predicate-Object (SPO). 
\begin{figure}[h]
\centering
\includegraphics[width=.4\textwidth]{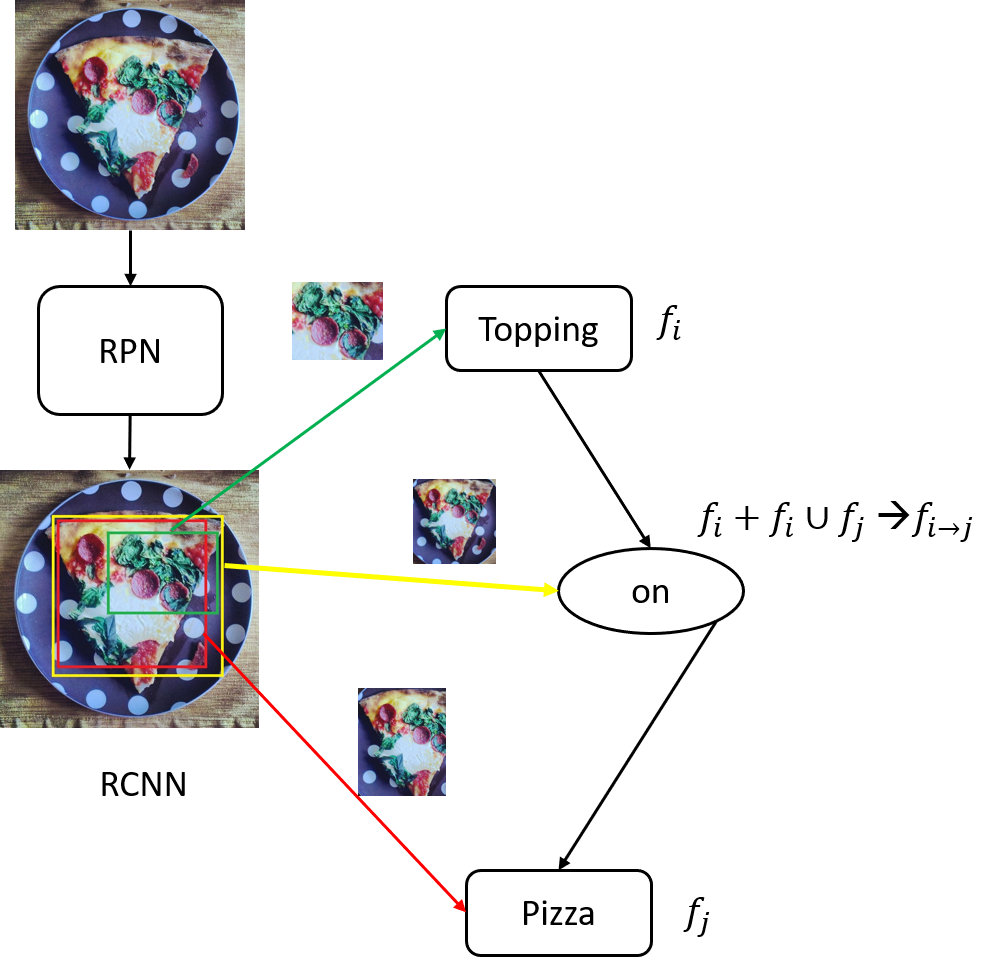}
\caption{Scene-Graph (SG) Generator Network Demonstrating The Link Between Object Features and Relationship Features. This Concept is Used in Scene-Graph Inference. Here, Instead of Relying on the Error-Prone Inference of Scene Graph Predictor Models, We Used Low Level Image Features Directly From RCNN from \cite{li2018factorizable}, But The Composition is Derived From Scene Graph Predictor.}
\label{fig:SG}
\end{figure}
The mathematical model of SG/SPO triplet generator $f_{SG}(.)$  and equations for prediction can denoted as the followings equations.
\begin{equation} \label{eq:chap3:fSG}
\begin{split}
 f_{SG}(\textbf{x}) & = \prod\limits_{x_j\in \textbf{x}}^{j\neq i} \mathrm{Pr}(x_{j \rightarrow i} \mid f_j, \text{ } f_{j \rightarrow i})  \\
 & \text{ } \text{ } \text{ } \prod\limits_{x_j\in \textbf{x}}^{j\neq i} \mathrm{Pr}(x_{i \rightarrow j} \mid f_i, \text{ } f_{i \rightarrow j}) \prod\limits_{x_i\in \textbf{x}}^{}  \mathrm{Pr}(x_i \mid f_i)  \text{ } 
\end{split} 
\end{equation}
where $x_{i \rightarrow j},x_{j \rightarrow i}, x_{i}, x_{j} \in \textbf{x}$ is the list of available objects and relationships for classification. Consideration of both $x_{i \rightarrow j}$ and $x_{j \rightarrow i}$ indicates that the context and union of the contexts are learned in both directions, to evolve the intuitions based on the spatial information. 
Our proposal for SPO embedding had taken the form of this following equation as the Scene Graph network tries to maximize the prediction of the individual objects. 
\begin{equation}
\{ \text{ } \mathrm{Pr}(x_i = s\mid S_{s})\text{ };\text{ }\mathrm{Pr}(x_i = p\mid S_{p})\text{ };\text{ }\mathrm{Pr}(x_i = o\mid S_{o}) \text{ }\}
\end{equation}
while we utilized the image feature embedding for subject $f_i = S_{s}$, predicate $f_{i \rightarrow j} = S_{p}$ and object $f_j = S_{o}$ as, 
\begin{equation}
    \{\text{ }S_{s}\text{ };\text{ }S_{p}\text{ };\text{ }S_{o}\text{ }\}
\end{equation}
However, this form is unstable and does not generalize the structural properties of graph and even extending the dimension of the representation three times higher. Hence, we define generalized structural properties TP$_{sgt}$R, where the positional information are product-ed with context.  
If we consider an image $\textbf{I}$ with regional visual features represented as $\{v_1,\ldots,v_N\}$, the SPO generator can learn to generate the relationships among various objects in the images. We can define the SPO generator as any model $f_{SG}(.)$ which derives the necessary information from the regional CNN infrastructure. Defining the SPO generator equation as, 
\begin{equation}
    \textbf{S} = f_{SG}(\textbf{I}) = f_{SG}(\{v_1,\ldots,v_N\})  
\end{equation}
we have defined the SG triplets as $S_i$ and $\{S_1,S_2,\ldots,S_n\} \in \textbf{S}$ are the scene graphs set with $n$ individuals in the set $\textbf{S}$ for an image $\textbf{I}$ and we have considered the Factorizable Net \cite{li2018factorizable} model for derivation of the scene-graphs triplets. Figure \ref{fig:SG} provided an overview of the Scene-Graph (SG) generator network.

For positioning, Tensor Product Scene-Graph-Triplet Representation (TP$_{sgt}$R) used a Hadamard matrix for derivation of the TPR graph representation $s_i$ from the SPO triplet $S_i$ denoted as the following,
\begin{equation}
    S_i = \{S_{s,i},S_{p,i},S_{o,i}\}
\end{equation}
and for each $S_i$, we have TP$_{sgt}$R $s_i$  and we have defined the whole set of TP$_{sgt}$R as $\textbf{s}$ with $\{s_1,s_2,\ldots,s_n\} \in \textbf{s}$ with $s_i$ and defined as the following equation. 
\begin{equation} \label{eqn:SPO_TPR}
    s_i = ( S_{s,i} r^T_1 + S_{p,i} r^T_2 + S_{o,i} r^T_3 )
\end{equation}
where we have defined $S_i$ as $\{S_{s,i},S_{p,i},S_{o,i}\}$ where each scene graph consists of an element S, P, and O as vector $S_{s,i}$, $S_{p,i}$, $S_{o,i}$ respectively. Here, $r^T_1,r^T_2,\ldots,r^T_3$ is derived from a $\{4\times 4\}$  Hadamard matrix normalized with 2-norm. We denote the $4\times 4$ Hadamard matrix as $\textbf{r}$ consisting of $\{r_0,r_1,r_2,r_4\}$ columns and normalized with 2-norm of column. 
For image captioning application, TP$_{sgt}$R helps in providing several discrete interaction information through the required graphical layer based representation interface and these can be used as neuro-symbolic description of a situation or scenario in an image. While, SPO content embedding can be defined in various ways, the most effective way can be when visual features are used directly for embedding as these can generate the maximum likelihood. On top of that, we have proposed to involve the positioning information into the system through a tensor product and create highly effective representation.

\begin{figure}[h]
\centering
\includegraphics[width=.5\textwidth]{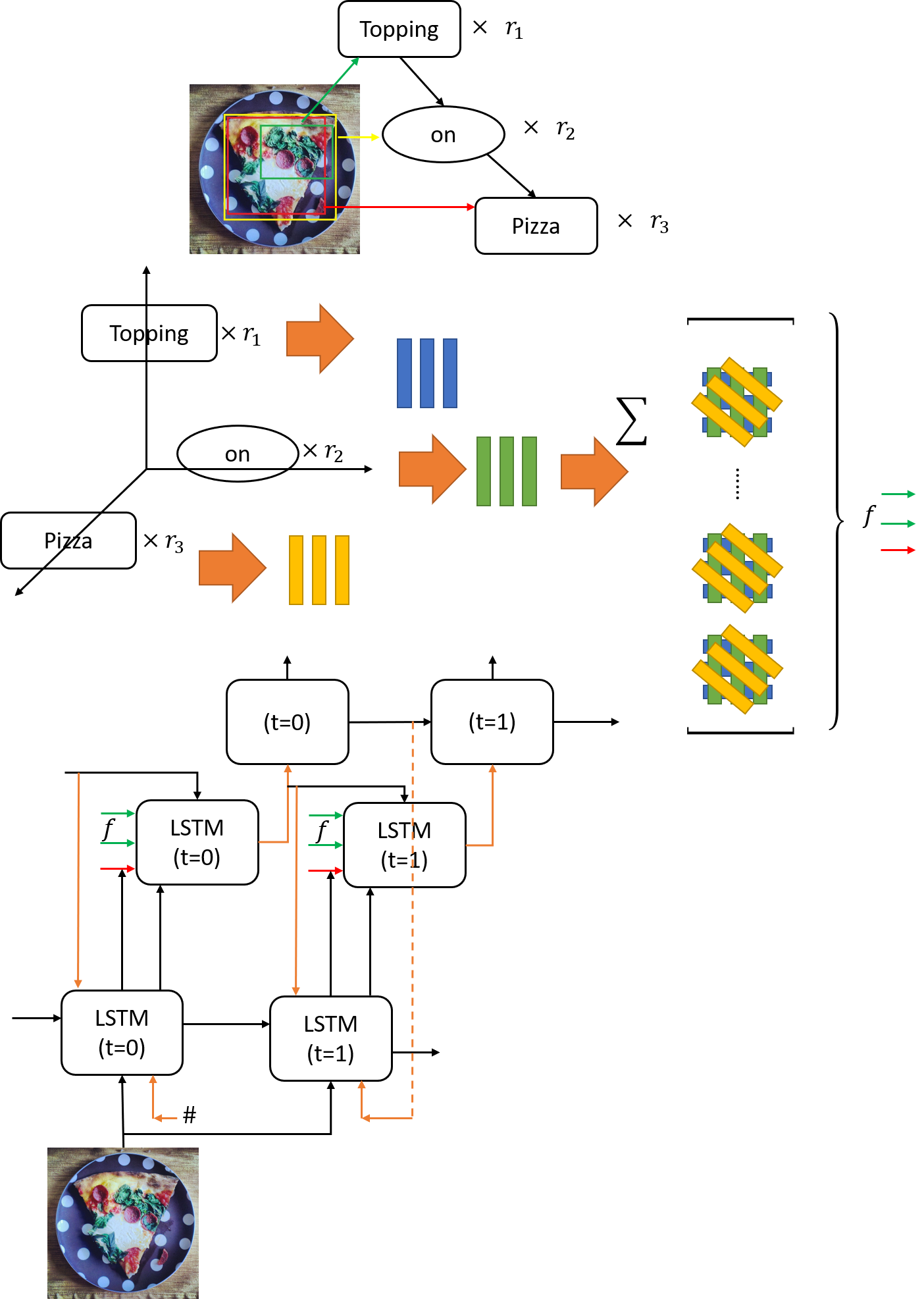}  
\caption{Architecture of Image Top-Down TP$_{sgt}$R Bottom-Up (TP$_{sgt}$R-TDBU).} 
\label{fig:SPO-TDBU}
\end{figure}

\subsection{Image Top-Down TP$_{sgt}$R Bottom-Up} 
Image Top-Down TP$_{sgt}$R Bottom-Up (TP$_{sgt}$R-TDBU) is inspired by the top-down-bottom-up approach \cite{anderson2018bottom}, but utilizes TP$_{sgt}$R features for inference and caption generation. Diagram of Image Top-Down TP$_{sgt}$R Bottom-Up (TP$_{sgt}$R-TDBU) in provided in Figure \ref{fig:SPO-TDBU}, where the language generator utilizes the novel graph embedding TP$_{sgt}$R for caption generation.  
Mathematically, Image Top-Down TP$_{sgt}$R Bottom-Up, denoted as $f_{itb}(.)$, can be described as the followings probability distribution estimation.
\begin{equation} \label{eq:itb}
\begin{split}
 f & _{itb}(\textbf{v}) = \prod\limits_{k}^{} \mathrm{Pr}(w_k \mid \textbf{x}, \text{ } \textbf{W}_{L_1}, \text{ }\textbf{W}_{L_2}) \prod\limits_{x_i\in \textbf{x}}^{} \mathrm{Pr}(S_i \mid f_i,  \text{ }\textbf{W}_1)  \\
 & = \prod\limits_{k}^{} \mathrm{Pr}(w_k \mid \{S_1,..,S_K\}, \textbf{W}_{L_1}, \textbf{W}_{L_2}) \prod\limits_{x_i\in \textbf{x}}^{} \mathrm{Pr}(S_i \mid f_i,  \textbf{W}_1)  \\
 & = \prod\limits_{k}^{} \mathrm{Pr}(w_k \mid \left( \frac{1}{K}\sum\limits_{m=1}^{K} f_m, \textbf{W}_{L_1} \right),  \left( \sum\limits_{m=1}^{N} \tilde{a}_ms_m, \textbf{W}_{L_2}\right) )  \\
 & \text{ }\text{ }\text{ } \prod\limits_{x_i\in \textbf{x}}^{} \mathrm{Pr}(S_i \mid f_i,  \text{ }\textbf{W}_1)  \\
 & = \prod\limits_{k}^{} \mathrm{Q}_{IC}(w_k \mid \frac{1}{K}\sum\limits_{m=1}^{K} f_m,   \sum\limits_{m=1}^{N} \tilde{a}_ms_m) \prod\limits_{x_i\in \textbf{x}}^{} \mathrm{Q}(S_i \mid f_i) 
\end{split} 
\end{equation}
using the weights of the two LSTM architectures and is denoted as $\textbf{W}_{L_1}$ and $\textbf{W}_{L_2}$, $w_i$ as words of sentences, $f_i$ as regional image features, $S_i$ as scene-graph triplets, $s_i$ as TP$_{sgt}$R, $\tilde{a}_ms_m$ as intermediate learnt parameters, $\mathrm{Q}_{IC}(.)$ and $\mathrm{Q}(.)$ are the Image Caption and Scene-Graph generator function respectively. $\mathrm{Q}(.)$ derives $\textbf{x}$ (Scene-Graph information) from $\textbf{v}$ of $\textbf{I}$.

We defined the Image Top-Down TP$_{sgt}$R Bottom-Up (TP$_{sgt}$R-TDBU) approach with top down features from the lower level ResNet CNN features and  TP$_{sgt}$R features as bottom refinement. Visual feature helps in providing a comprehensive overview of the image that cannot be summarized by the weighted summation of the regional representation, however, since the regional features are sparse and individually comprehensive, they sometime provide better overview and create use-able compositions. 
\begin{equation}
    \overline{v} = v
\end{equation} 
where we have $v$ as the ResNet image feature and $\overline{v}$ is initiated with $v$.
This $\overline{v}$ feature of the whole scenario provides a better overview of the image and the different components. The feature for the first recurrent unit $LSTM_1(.)$ for Top Down is denoted as $\textbf{x}^1_{t}$.
\begin{equation}
    \textbf{x}^1_{t} = [\textbf{h}^2_{t-1}; \overline{v}; \textbf{W}_{e}\textbf{x}^2_{t-1}]
\end{equation}
\begin{equation}
    \textbf{h}^1_{t} = LSTM_1(\textbf{x}^1_{t}, \textbf{h}^1_{t-1})
\end{equation}
Next, we have calculate the intermediate weights for regional attention, and here, the attention is regional TP$_{sgt}$R attention and is denoted as $\hat{s}$. 
\begin{equation}
    a_{i,t} = \textbf{W}_a \tanh (\textbf{W}_b\textbf{s}_i + \textbf{W}_b\textbf{h}^1_{t})
\end{equation}
\begin{equation}
   \tilde{\textbf{a}} = \text{softmax} (\textbf{a})
\end{equation}
\begin{equation}
    \hat{s} = \sum\limits_{i=1}^{{N_s}} s_i\tilde{a_i}
\end{equation}
where $s_i$ is derived from Equation \ref{eqn:SPO_TPR} and is utilized instead of image features. Here we used the Scene-Graph based SPO positioning TPR (TP$_{sgt}$R) from Equation \ref{eqn:SPO_TPR} as the feature representing the whole activity representation and their dependencies on spatially external attributes instead of the low level regional CNN feature characteristics. The language decoder $LSTM_2(.)$  works more for joint fusion of features with the following representation $\textbf{x}^2_{t}$ as input.
\begin{equation}
    \textbf{x}^2_{t} = [\textbf{h}^1_{t};\hat{s}]
\end{equation}
\begin{equation}
    \textbf{h}^2_{t} = LSTM_2(\textbf{x}^2_{t}, \textbf{h}^2_{t-1})
\end{equation}
where we generate the maximum likelihood of $(\textbf{W}_{hx} \textbf{h}^2_{t})$ for determination of the words as sentences. This architecture is the baseline for our TP$_{sgt}$R features and achieved 33.8 BLEU\_4 evaluation. 

\subsection{Semantic Top-Down TP$_{sgt}$R Bottom-Up}
In Semantic Top-Down TP$_{sgt}$R Bottom-Up (TP$_{sgt}$R-sTDBU), the most important characteristic of this approach is that we tried not to involve the whole image data, as it is defined in some totally different feature space and also it tends to manipulate the language decoder to specific direction based on the abstract visibility of the transformation, learnt by the model. Figure \ref{fig:SPO-sTDBU} provided a descriptive diagram of the SPO-sTDBU architecture at the feature level. Also, mathematically, Semantic Top-Down TP$_{sgt}$R Bottom-Up (denoted as $f_{stb}(.)$) can be described as the followings set of probability distribution estimation using the weights of the two LSTM architectures and is denoted as $\textbf{W}_{L_1}$ and $\textbf{W}_{L_2}$. 
\begin{equation}
\begin{split}
 f & _{stb}(\textbf{v}) = \prod\limits_{k}^{} \mathrm{Pr}(w_k \mid \textbf{x}, \text{ } \textbf{W}_{L_1}, \text{ }\textbf{W}_{L_2}) \prod\limits_{x_i\in \textbf{x}}^{} \mathrm{Pr}(S_i \mid f_i,  \text{ }\textbf{W}_1)  \\
 & = \prod\limits_{k}^{} \mathrm{Pr}(w_k \mid \{S_1,..,S_K\}, \textbf{W}_{L_1}, \textbf{W}_{L_2}) \prod\limits_{x_i\in \textbf{x}}^{} \mathrm{Pr}(S_i \mid f_i,  \textbf{W}_1)  \\
 & = \prod\limits_{k}^{} \mathrm{Pr}(w_k \mid \left( \frac{1}{K}\sum\limits_{m=1}^{K} S_m, \textbf{W}_{L_1} \right),\textbf{s}_e, \left( \sum\limits_{m=1}^{N} \tilde{a}_ms_m, \textbf{W}_{L_2}\right) )  \\
 & \text{ }\text{ }\text{ } \prod\limits_{x_i\in \textbf{x}}^{} \mathrm{Pr}(S_i \mid f_i,  \text{ }\textbf{W}_1)  \\
 & = \prod\limits_{k}^{} \mathrm{Q}_{IC}(w_k \mid \frac{1}{K}\sum\limits_{m=1}^{K} S_m,   \sum\limits_{m=1}^{N} \tilde{a}_ms_m) \prod\limits_{x_i\in \textbf{x}}^{} \mathrm{Q}(S_i \mid f_i) 
\end{split} 
\end{equation}
where $\textbf{s}_e$ is the Semantic information. 
This caption generator architecture works with Semantic Top-Down, while the Bottom-Up operates with the TP$_{sgt}$R features. The previous architecture \cite{anderson2018bottom} operated with regional object representation throughout, we argue that the descriptive operation representations and semantics compose the best features at the intermediate level of the architecture. 
We defined the Semantic Top-Down TP$_{sgt}$R Bottom-Up (TP$_{sgt}$R-sTDBU) approach with Top-Down features gathering the overall semantic of the image features from the SPO embedding (here Regional-CNN) features and the Bottom-Up selects from the set of detailed description in the form TP$_{sgt}$R. Here, $\overline{v}_s$ initialization can be defined by the following equations. 
\begin{equation}
    \overline{v}_s = \sum \limits_{i=1}^{k} [S_{s,i}; S_{p,i}; S_{o,i}]
\end{equation} 
This $\overline{v}$ feature of the whole scenario provides a better overview of the image than the transformed visual features. It gathers the overall activities of the different components in the form of TP$_{sgt}$R components and is not based on the absolute image. The first recurrent unit can be defined as $LSTM_1(.)$ with inputs as $\textbf{x}^1_{t}$. 
\begin{equation}
    \textbf{x}^1_{t} = [\textbf{h}^2_{t-1}; \overline{v}_s; \textbf{W}_{e}\textbf{x}^2_{t-1}]
\end{equation}
\begin{equation}
    \textbf{h}^1_{t} = LSTM_1(\textbf{x}^1_{t}, \textbf{h}^1_{t-1})
\end{equation} 
The TP$_{sgt}$R is described with Equation \ref{eqn:SPO_TPR} for each $s_i$ of $\{s_1,\ldots,s_k\} \in \textbf{s}$  where the whole set of TP$_{sgt}$R is denoted as $\textbf{s}$. 
Next, the intermediate selection of the weighted activities and selection of semantic can be represented as the following equations. 
\begin{equation}
    a_{i,t} = \textbf{W}_a \tanh (\textbf{W}_b\textbf{s}_i + \textbf{W}_c\textbf{h}^1_{t})
\end{equation}
\begin{equation}
   \tilde{\textbf{a}} = \text{softmax} (\textbf{a})
\end{equation}
\begin{equation}
    \hat{s} = \sum\limits_{i=1}^{{N_s}} s_i\tilde{a_i}
\end{equation}
where TP$_{sgt}$R $s_i$ is derived from Equation \ref{eqn:SPO_TPR} and is utilized instead of visual representations from image based regional CNN feature characteristics. 
Here, we use the Scene-Graph based TP$_{sgt}$R, which is a positional transformed embedding of the feature space, as the main input and is characterized by the activities of the objects with its surrounding instead of the objects themselves. Next, we used Semantics decomposition of the Top-Down features with $\textbf{s}_e$ as the followings,
\begin{equation}
    \textbf{h}^{1_n}_{t} = \textbf{W}_{S_1} \textbf{s}_e \odot \textbf{W}_{1_n} \textbf{h}^1_{t} 
\end{equation}
\begin{equation}
    \textbf{x}^2_{t} = [\textbf{h}^{1_n}_{t}; \hat{s}]
\end{equation}
\begin{equation}
    \textbf{h}^{2_n}_{t} = \textbf{W}_{S_2} \textbf{s}_e \odot \textbf{W}_{2_n} \textbf{h}^2_{t-1} 
\end{equation}
\begin{equation}
    \textbf{h}^2_{t} = LSTM_2(\textbf{x}^2_{t}, \textbf{h}^{2_n}_{t})
\end{equation}
where the maximum likelihood of $(\textbf{W}_{hx} \textbf{h}^2_{t})$ determines the words as sentences for Semantic Top-Down TP$_{sgt}$R Bottom-Up (TP$_{sgt}$R-sTDBU).

\section{Result \& Analysis}\label{sec:res}

\begin{table*}[!h]
\centering
\caption{Performance Comparison \& Analysis for Different Image Captioning Architectures for Different Metrics}
\begin{tabular}{|c|c|c|c|c|c|c|c|c|}
\hline
 Algorithm & CIDEr-D & Bleu\_4 & Bleu\_3 & Bleu\_2 & Bleu\_1 & RG\_L & METEOR  &  SPICE \\ 
\hline \hline
    Adaptive \cite{lu2017knowing}  & 1.085 & 0.332 & 0.439 & 0.580 & 0.742 & -- & 0.266 & --  \\
    MSM \cite{yao2017boosting}  & 0.986 & 0.325 & 0.429 & 0.565 & 0.730 & -- & 0.251 & --  \\ 
    Att2in \cite{rennie2017self}  & 1.01 & 0.313 & -- & -- & -- & -- & 0.260 & --  \\  
    Top-Down  \cite{anderson2018bottom}  & 1.135 & 0.362 & -- & -- & 0.772 & 0.564 & 0.270 & 0.203  \\ %
     TPGN \cite{huang2018tensor}  & 0.909 & 0.305 & 0.406 & 0.539 & 0.709 & -- & 0.243 & --  \\ %
     ATPL \cite{huang2019tensor}  & 1.013 & 0.335 & 0.437 & 0.572 & 0.733 & -- & 0.258 & --  \\ %
    Attribute-Attention \cite{chen2018show} & 1.044 & 0.338 & 0.443 & 0.579 & 0.743 & 0.549  & -- & -- \\ %
    SCN \cite{gan2017semantic} & 1.012 & 0.330 & 0.433 & 0.566 & 0.728 & -- & 0.257 & --  \\ 
    (Semantic TPR) \cite{chiranjib2019semantic}  & {1.022} & {0.338} & {0.443} & {0.578} & {0.737} & {0.546} & {0.256} & {0.1862}  \\ 
    \hline\hline
 
 Image Top-Down & & & & & & & & \\  						
    TP$_{sgt}$R Bottom-Up   & 1.017 & 0.338 & 0.444 & 0.581 & 0.737 & 0.546 & 0.255 & 0.186 \\
    \textbf{Semantic Top-Down} & & & & & & & & \\
    \textbf{TP$_{sgt}$R Bottom-Up} & \textbf{1.082} & \textbf{0.349} & \textbf{0.456} & \textbf{0.594} & \textbf{0.752} & \textbf{0.554} & \textbf{0.265} & \textbf{0.196} \\  
\hline
\end{tabular}
\label{table:table1}
\end{table*}

\subsection{Dataset Preparation \& Training}
MSCOCO data is used for experiments. It consists of 123K  images and 566K sentences for training and each image is associated with at least five sentences from a vocabulary of 8791 words with 5K images (with 25K sentences) for validation and 5K images (with 25K sentences) for testing \cite{gan2017semantic,Karpathy2014Deep}. ResNet features ($\textbf{v}$ with 2048 dimension), Tag features ($\textbf{s}_e$ with 999 dimension) \cite{gan2017semantic} and TP$_{sgt}$R  using \cite{li2018factorizable} model consisted of the different representation used for experiments. Tag contributes more when used with other correlated features and the correlation based fusion has been the turning point for these image captioning application. 

\subsection{Results Analysis}
\begin{figure*}[h]
\centering
\includegraphics[width=\textwidth]{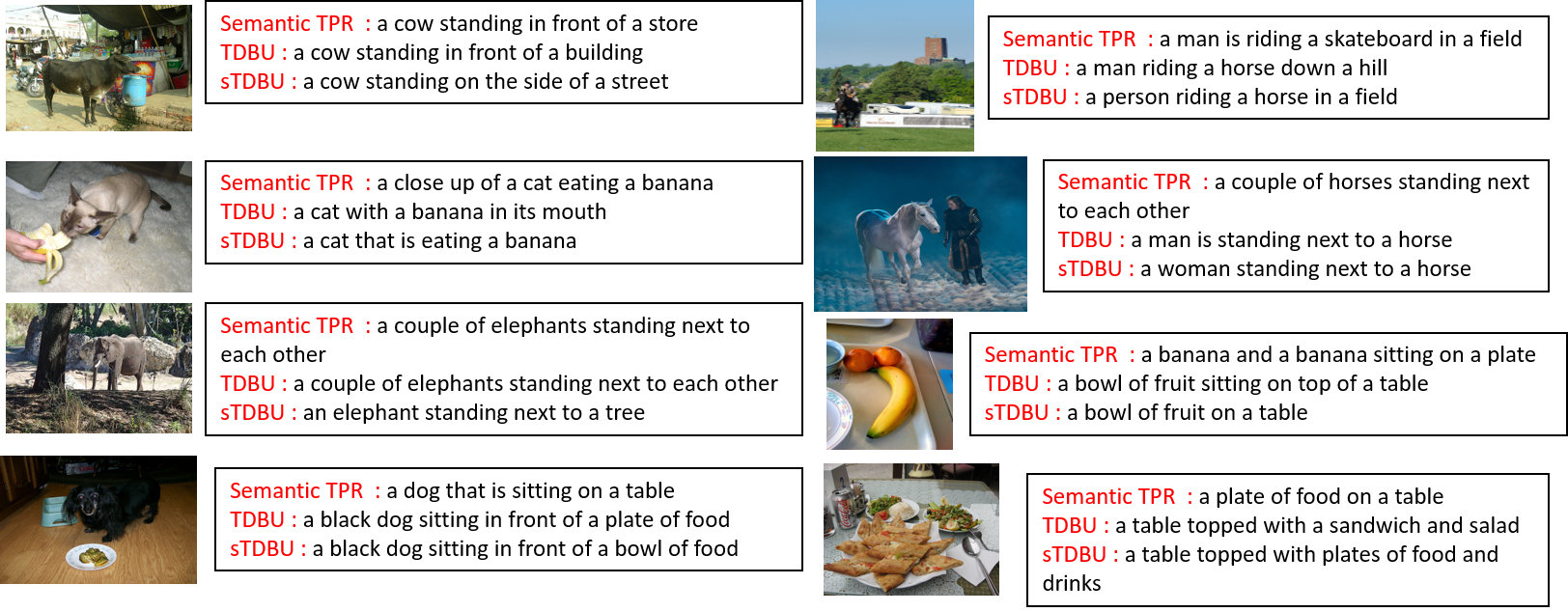}  
\caption{Qualitative Analysis of the Image Captions Generated from TP$_{sgt}$R-sTDBU,  TP$_{sgt}$R-TDBU and Semantic TPR \cite{chiranjib2019semantic}.} \label{fig:QualitativeAnalysis1}
\end{figure*}

Different language metrics like Bleu\_n $(n=1,2,3,4)$, METEOR, ROUGE\_L, CIDEr-D and SPICE are provided as these are standardized in the community and are used for performance evaluation and comparison of our model. However, each reflect very limited perspective of the generated captions and hence Figure \ref{fig:QualitativeAnalysis1} provided a qualitative comparison of the generated captions for our new architecture. 
Table \ref{table:table1} provided a comparative study of our models with some of the existing state-of-the-art works in the image captioning domain using image features and also with others like Attribute-Attention \cite{chen2018show}, RCNN \cite{lu2018neural, anderson2018bottom}, where \cite{anderson2018bottom} used advanced features for their work and yet our work is comparable. 
The main functional characteristics of our work is the tensor product based positional aware representation, which embed the structural characteristics of objects and interactions in a graph.  Generated/trained structural features undergo approximation and then compose new representation through different circumstances of weights, where the weights never been able to represent the whole dataset with different characteristics. Most of our new architectures performed very well and either outperformed or at least same with the existing architectures, which do not have concrete reasoning behind their working principles.
With the introduction of semantics, our TP$_{sgt}$R-sTDBU architecture performed much better than the existing works and performed better than the TP$_{sgt}$R-TDBU architecture, establishing the fact that they generate better strategy for caption generation. However, the TP$_{sgt}$ features and the Scene-Graph feature generator was trained with images which are not correlated to MSCOCO dataset and the accuracy of detection is around $~$28\%. Also, the average number of Scene-Graph features detected is 8-9 on average and maximum at 15, which is way less than \cite{anderson2018bottom}, where they extracted 36 regional features and did not disclose the principle of selection of these regional features. Hence, in comparison to \cite{anderson2018bottom}, which achieved 36.2\% BLEU\_4, our TP$_{sgt}$R-sTDBU architecture  achieved 34.9\% BLEU\_4 and is still comparable because of the fact that the Scene-Graph generator model \cite{li2018factorizable} was totally trained with a different data.  Though statistical metrics provide many qualitative insights of the generated languages, they hardly reflected any language attribute quality related to meaning, grammar, correct part-of-speech etc. These can only be evaluated through reading and hence we provided comparison in terms of diversity and descriptive attributes in Figure \ref{fig:QualitativeAnalysis1}. 

\section{Conclusion} \label{section:discussion}
In this work, we discussed some improvements of the existing structures of feature structuring definition and demonstrated that our approaches are far better than previous works in all the possible metrics both theoretically and experimentally. We introduced TP$_{sgt}$R for graph embedding and its interaction with other informative structures of the memory network and leveraged the structural variations it can generate for identification of attributes and interaction in images to appear in the form of a sentences. Our mission was to generate better representation and something that can be generalized and scaled for large applications related to media and aid non-human understand structures and images. While, TP$_{sgt}$R succeeded in gathering improvement, we introduced different feature fusion approaches which are as good as some of the state-of-the-art features. The future works can be directed to more sophistication of the structures and diversification in quality and introduction of other useful components of the data that differentiate  and generalize the attributes to its unique counterparts.





%




\bibliographystyle{IEEEtran}
\bibliography{spo_paper}

\end{document}